\documentclass{ieeeaccess}
\usepackage{cite}
\usepackage{amsmath,amssymb,amsfonts}
\usepackage{algorithmic}
\usepackage{graphicx}
\usepackage{textcomp}
\usepackage{url}
\usepackage[shortcuts]{extdash} 
\usepackage{color,array}
\usepackage{adjustbox}
\usepackage{comment}
\usepackage{multirow}
\usepackage{subfig}
\usepackage{caption}
\usepackage{cuted}
\usepackage{amsmath}
\usepackage[T1]{fontenc}
\usepackage{tgbonum}
\usepackage{placeins}
\usepackage{cite}
\usepackage{hyperref}
\usepackage{multicol,lipsum}
\usepackage[splitrule]{footmisc}

\renewcommand{\footnoterule}{\vfill\kern -3pt \hrule width 0.4\columnwidth \kern 2.6pt}
\usepackage{etoolbox}
\AfterEndEnvironment{strip}{\leavevmode}
\usepackage{lipsum}
\renewcommand{\footnoterule}{\vfill\kern -3pt \hrule width 0.4\columnwidth \kern 2.6pt}

\newcommand{\orcid}[1]{\href{https://orcid.org/#1}{\includegraphics[width=8pt]{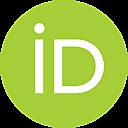}}}

\definecolor{babyblue}{rgb}{0.54, 0.81, 0.94}
\definecolor{almond}{rgb}{0.94, 0.87, 0.8}
\definecolor{cambridgeblue}{rgb}{0.64, 0.76, 0.68}
\definecolor{mauvelous}{rgb}{0.94, 0.6, 0.67}
\definecolor{upsdellred}{rgb}{0.90, 0.09, 0.13}
\definecolor{darkspringgreen}{rgb}{0.09, 0.45, 0.27}
\definecolor{oceanboatblue}{rgb}{0.0, 0.47, 0.75}

\definecolor{backnvbG}{RGB}{65, 105, 225}
\definecolor{backorange}{RGB}{255, 165, 0}
\definecolor{backgreen}{RGB}{80, 220, 71}
\definecolor{backred}{RGB}{250,70,51}

\definecolor{backgG}{RGB}{255, 255, 153}
\definecolor{tagtxtG}{RGB}{102, 102, 0}
\definecolor{backgPc}{RGB}{179, 255, 179}
\definecolor{tagtxtPc}{RGB}{0, 102, 0}
\definecolor{backgPw}{RGB}{255, 179, 179}
\definecolor{backgPw}{rgb}{0.0, 1.0, 1.0}
\definecolor{tagtxtPw}{RGB}{0.0, 1.0, 1.0}
\definecolor{backgPo}{rgb}{0.0, 1.0, 1.0}
\definecolor{tagtxtPo}{RGB}{102, 0, 0}
\definecolor{backgPm}{rgb}{0.98, 0.81, 0.69}
\definecolor{tagtxtPm}{RGB}{0,1,1}

\DeclareMathOperator{\BERT}{BERT}

\renewcommand{\vec}[1]{\boldsymbol{\mathbf{#1}}}

\usepackage[spaces,hyphens]{xurl}
\def\BibTeX{{\rm B\kern-.05em{\sc i\kern-.025em b}\kern-.08em
    T\kern-.1667em\lower.7ex\hbox{E}\kern-.125emX}}
\begin{document}

\history{Received 25 May 2023, accepted 16 June 2023, date of publication 4 July 2023, date of current version 30 August 2023.}
\doi{10.1109/ACCESS.2023.3292300}

\title{Generation of Highlights from Research Papers Using Pointer-Generator Networks and SciBERT Embeddings}
\author{\uppercase{Tohida Rehman} \orcid{0000-0002-3578-1316} \authorrefmark{1} , 
\uppercase{Debarshi Kumar Sanyal \orcid{0000-0001-8723-5002}\authorrefmark{2}, Samiran Chattopadhyay \orcid{0000-0002-8929-9605}\authorrefmark{1,3}, Plaban Kumar Bhowmick \orcid{0000-0002-6573-0093}\authorrefmark{4}, and Partha Pratim Das} \orcid{0000-0003-1435-6051}\authorrefmark{5,6},(Member, IEEE)}
\address[1]{Department of Information Technology, Jadavpur University, Salt Lake Campus, Kolkata, West Bengal 700106, India}
\address[2]{School of Mathematical and Computational Sciences, Indian Association for the Cultivation of Science, Kolkata, West Bengal 700032, India}
\address[3]{Techno India University, West Bengal, Kolkata 700091, India}
\address[4]{G S Sanyal School of Telecommunication, Indian Institute of Technology Kharagpur, Kharagpur, West Bengal 721302, India}
\address[5]{Department of Computer Science and Engineering, Indian Institute of Technology Kharagpur, Kharagpur, West Bengal 721302, India}
\address[6]{Department of Computer Science, Ashoka University, Sonipat, Haryana 131029, India}

\markboth
{Rehman  \headeretal: Generation of Highlights from Research Papers Using Pointer-Generator Networks and SciBERT Embeddings}
{Rehman \headeretal: Generation of Highlights from Research Papers Using Pointer-Generator Networks and SciBERT Embeddings}

\corresp{Corresponding author: Tohida Rehman (e-mail: {\tt{tohidarehman.it@jadavpuruniversity.in})}}

\begin{abstract}
Nowadays many research articles are prefaced with research highlights to  summarize the main findings of the paper. Highlights not only help researchers precisely and quickly identify the contributions of a paper, they also enhance the discoverability of the article via search engines. 
We aim to automatically construct research highlights given certain segments of a research paper. We use a  pointer-generator network with coverage mechanism and a contextual embedding layer at the input that encodes the input tokens into SciBERT embeddings. We test our model on a benchmark dataset, CSPubSum, and also present MixSub, a new multi-disciplinary corpus of papers for automatic research highlight generation. For both CSPubSum and MixSub, we have observed that the proposed model achieves the best performance compared to related variants and other models proposed in the literature. On the CSPubSum dataset, our model achieves the best performance when the input is only the abstract of a paper as opposed to other segments of the paper. It produces ROUGE-1, ROUGE-2 and ROUGE-L F1-scores of 38.26, 14.26 and 35.51, respectively, METEOR score of 32.62, and BERTScore F1 of 86.65 which outperform all other baselines. On the new MixSub dataset, where only the abstract is the input, our proposed model (when trained on the whole training corpus without distinguishing between the subject categories) achieves ROUGE-1, ROUGE-2 and ROUGE-L F1-scores of 31.78, 9.76 and 29.3, respectively, METEOR score of 24.00, and BERTScore F1 of 85.25.
\end{abstract}


\titlepgskip=-15pt

\maketitle
\section{Introduction}\label{sec1}

Scientific publications are growing at an exponential rate \cite{bornmann2021growth}. It has been reported that the number of scientific articles doubles roughly every nine years \cite{van2014global}. Even in a limited sub-field, scientists find it very challenging to keep track of the cutting edge of research. Therefore, to make it easier for researchers to appreciate the main import of a paper, publishers have adopted many novel presentation techniques. One recent trend is to complement the abstract of a paper with \textit{research highlights}, a list of points summarizing the main findings of the paper. Research highlights are typically written by the author along with the abstract. They are often easier to read and grasp than a longer paragraph, especially on hand-held devices. Moreover, research highlights can be used by search engines for indexing the articles and subsequently, retrieve or recommend them to the appropriate users. Yet, not all scholarly articles contain research highlights written by the authors.

Research highlights and abstract are both \textit{summaries} of the research paper. 
Text summarization is a process to present the gist of a source document or a set of related documents. The main benefit of text summarization is that it reduces the amount of time the reader has to spend to extract the main information in the document. Extractive summarization and abstractive summarization are two broad approaches used in automatic text summarization \cite{luhn1958automatic, 9623462}. Extractive approaches \cite{kupiec1995trainable} simply copy some relevant sentences from the documents and ignore the rest. Abstractive approaches \cite{el2021automatic} can induce new relevant words in the summary in the same way that a person does -- they first read the entire text, comprehend it, and then summarize using suitable new words. Therefore, abstractive approaches typically provide better summaries compared to those produced by extractive methods. 

In this paper, we aim to extract research highlights from a research paper using abstractive approaches. From a simple manual analysis, we found that most of the information present in research highlights occur in the abstract, introduction, and conclusion sections of a paper. Therefore, we provide these sections and their combinations as inputs to our summarizer. Our model is an adaptation of the pointer-generator network with coverage mechanism \cite{See2017GetTT}. However, unlike the original model, we use an additional embedding layer at the input. This layer  encodes each word of the input document with embeddings from SciBERT, which is a BERT model trained on a large corpus of scientific documents. We expect these contextual embeddings to help the model generate better quality abstractive summary compared to that produced by the vanilla model.

The main contributions of this paper are: 
\begin{enumerate}
    \item  Our method automatically generates research highlights from a scientific research paper. We propose a technique to combine a SciBERT \cite{beltagy-etal-2019-scibert} pre-trained layer of word embeddings with a pointer-generator network that also uses a coverage mechanism.  To the best of our knowledge, this work is the first attempt to use SciBERT with a pointer-generator model augmented with coverage mechanism \cite{See2017GetTT} to generate research highlights. 
    \item We present a new multi-disciplinary dataset named MixSub that contains research papers (with author-written highlights) from different subject domains.
    \item For one of the datasets, namely, the CSPubSum dataset, we analyze the performance of  generating research highlights for the following different input types: (a) the input is the abstract only, (b) the input is the introduction only, (c) the input is the conclusion only,  (d) the input comprises the abstract and the conclusion, (e) the  input comprises the introduction and the conclusion. For the MixSub dataset, we use only the abstract as the input. We use ROUGH \cite{lin2004rouge}, METEOR \cite{banerjee2005meteor}, and BERTScore \cite{zhang2019bertscore} metrics to evaluate the performance of the models. We show that our model performs better than existing baselines proposed for this task.
\end{enumerate}
The rest of this article is organized as follows. Section \ref{sec2} is an overview of prior work in the field. Section \ref{Methodology} describes the proposed model. Section \ref{Experiment} describes the experimental setup, the datasets we used, including a new dataset called MixSub which we construct as part of this work. Section \ref{Results} reports the results we obtained by using the CSPubSum dataset and our new MixSub dataset. Within this section, we compare our method with a few competitive pre-trained models in the literature, and analyze the energy consumption of the various models. Section \ref{Case Studies} depicts several case studies. Finally, the paper concludes in Section \ref{Conclusion}.

\section{Literature review}
\label{sec2}
The advancement of sequence-to-sequence models \cite{sutskever2014sequence} has significantly improved the state-of-the-art in abstractive summarization \cite{9328413}. Attention-based encoder of  with a beam-search decoder has achieved significant performance in  abstractive text summarization on DUC 2004 dataset \cite{bahdanau2014neural}. Convolutional attention-based conditional recurrent neural network was used to further improve the performance on Gigaword Corpus and DUC 2004 dataset \cite{chopra2016abstractive}. Nallapati et al. \cite{nallapati2016abstractive} proposed a model for abstractive text summarization based on attentional encoder decoder recurrent neural networks. To remove out-of-vocabulary words (OOV) and repeating words, a hybrid approach called pointer-generator network with coverage mechanism has been proposed \cite{See2017GetTT}. It can copy words from the source text by pointing and uses coverage to keep track of what is summarized to avoid repetition. To represent the semantic information of words more correctly, Anh and Trang \cite{anh2019abstractive} have used two pre-trained word embeddings, namely, word2vec and FastText, with a pointer-generator model for the CNN/Daily Mail dataset and achieved an impressive  performance. Du et al. \cite{9139935} proposed a model to extract summary based on fuzzy logic rules, multi-feature set and genetic algorithm on DUC2002 dataset.

Recently, pre-trained language models that generate contextual embeddings have become extremely popular and shown to achieve state-of-the-art results in many NLP tasks. Their mode of operation is as follows: train the model on a large corpus and then fine-tune it on various downstream task in NLP. Radford et al. \cite{radford2018improving} proposed Generative Pre-Training (GPT), which combines unsupervised pre-training and supervised fine-tuning, to improve language understanding. The implementation of the transformer architecture and its bidirectional encoder model BERT resulted in improved performance in downstream NLP tasks including text summarization \cite{devlin2018bert}. BERT trains a deep bidirectional transformer encoder, which learns interactions between left and right context, using a masked language modelling objective \cite{devlin2018bert}. For a new corpus, BERT can be fine-tuned for sentence-label and token-label tasks. Knowledge graphs (KGs) can be combined with BERT to capture the lexical, syntactic, and knowledge information at the same time \cite{zhang-etal-2019-ernie}. Researchers have also made available a few large deep neural models that are pre-trained specifically for the summarization task. A pre-trained model PEGASUS, trained using large pre-training corpora and a gap sentence generating task has been evaluated on 12 downstream summarization tasks \cite{zhang2020pegasus}. To overcome the disadvantages of limited input size in a BERT-based architecture, the BERT windowing method can be used \cite{aksenov2020abstractive}. Raffel et al.\cite{10.5555/3455716.3455856} proposed T5, which is an encoder-decoder model pre-trained on a multi-task mixture of unsupervised and supervised workloads, with each task transformed to text-to-text processing. BART \cite{lewis-etal-2020-bart} is a transformer encoder-encoder (seq2seq) model with a bidirectional BERT encoder and an autoregressive decoder (more specifically, generative pre-trained transformer or GPT). The pre-training task in BART entails changing the sequence of the original phrases at random and using a novel in-filling strategy that replaces text spans with a single mask token. BART is especially effective when fine-tuned for text generation, but it also performs well for comprehension tasks. 

Early work on extractive summarization of scientific documents was done with limited datasets, such as one of 188 document and summary pairs \cite{kupiec1995trainable} where all the documents were gathered from 21 scientific/technical publications.  A summarizing technique that focuses on the rhetorical status of assertions in 80 scientific articles, part of a larger corpus of 260 articles, has been developed by Teufel and Moens \cite{teufel-moens-2002-articles}. A sentence-based automatic summarizing system has been built based on feature extraction and query-focused methods \cite{visser2007sentence}. Lloret et al. \cite{lloret2013compendium} have proposed a method to automatically generate the abstract of a research paper in the biomedical domain. They used two approaches -- extractive as well as abstractive. 
To better deal with the long text of a research paper in abstractive summarization, a multiple timescale model of the gated recurrent unit (MTGRU) has been used in \cite{kim-etal-2016-towards}. They have contributed a new corpus containing pairs of (introduction, abstract) of computer science  papers from \url{arXiv.org}. Souza et al. \cite{9892526} have proposed a multi-view extractive text summarization approach for long scientific texts. 

Recent advancements have attempted to summarize entire research papers, focusing specifically on the generation of the paper title from the abstract (title-gen) and the generation of the abstract from the body of the paper (abstract-gen) in  biomedical domain \cite{nikolov2018data}. Since the keyphrases in a paper may be assumed to capture the main aspects of the paper, extraction \cite{santosh2020dake,santosh2020sasake} or generation \cite{santosh2021gazetteer,santosh2021hicova} of keyphrases is a related area of research. However, we do not discuss it further as it does not produce sentential forms typically observed in a summary.
  
\textit{Generating research highlights} from scientific articles is different than document summarization. Collins,  Augenstein and Riedel  \cite{collins2017supervised} have developed supervised machine learning methods to identify relevant highlights from the full text of a paper using a binary classifier. They also contributed a new benchmark dataset of URLs, which includes approximately 10,000 articles from computer science domain, labelled with relevant author-written highlights. Using multivariate regression methods for the same problem, Cagliero and Quatra in  \cite{cagliero2020extracting} selected the top-$k$ most relevant sentences from a paper as research highlights, unlike a simple binary classification of sentences as highlights or not. Note that this is also extractive in nature.  Rehman et al. \cite{rehman2021automatic} proposed an abstractive summarization model based on pointer-generator network with coverage and GloVe embeddings to generate research highlights from abstracts. Later, Rehman et al. \cite{rehman-etal-2022-named} combined named entity recognition with pointer-generator networks to improve the performance of their method. 
In contrast to previous works, in the current one we use pre-trained SciBERT word embeddings and propose a new dataset. 

\section{Methodology}
\label{Methodology}
We use pointer-generator networks to produce highlights from research papers. It consists of a seq2seq model with a BiLSTM encoder and an LSTM decoder with attention \cite{nallapati2016abstractive}. However, while the original model proposed by See, Liu, and Manning \cite{See2017GetTT} uses word-embeddings -- they are learned from scratch during training -- we use a pre-trained transformer to generate the contextual embeddings of the tokens in the input document. The architecture of our model is shown in Figure \ref{fig:model_diagram_scibert}. 
\Figure[t!](topskip=0pt, botskip=0pt, midskip=0pt) [width=.88\textwidth]{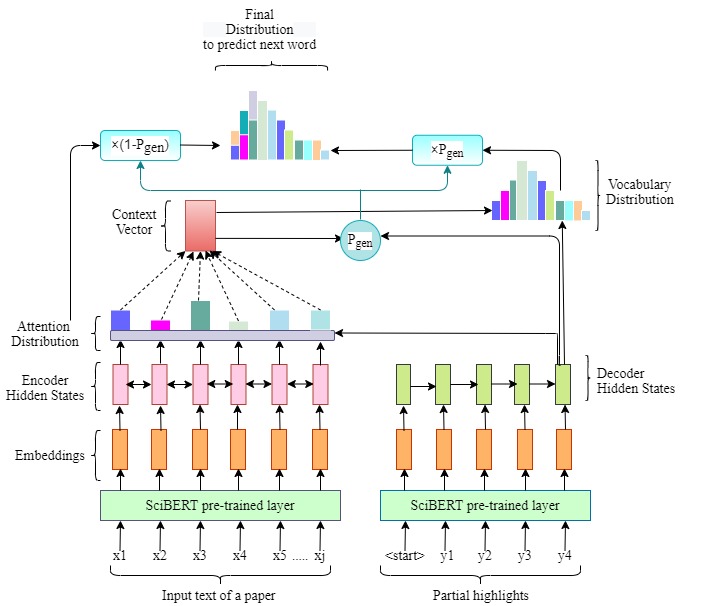}
{Proposed model: Pointer-generator network with coverage mechanism and SciBERT word embeddings.\label{fig:model_diagram_scibert}}

We perform experiments with 4 variants: (1) the original pointer-generator model proposed in \cite{See2017GetTT}, (2) pointer-generator model integrating coverage mechanism (proposed in \cite{tu2016modeling}), described in the same work \cite{See2017GetTT}, (3) pointer-generator model with SciBERT, and (4) pointer-generator model with SciBERT and coverage mechanism.
\subsection{BERT and SciBERT}
The pre-trained language model BERT stands for \textit{bidirectional encoder representations from transformers}. BERT has been pre-trained on the tasks of masked language modelling (MLM) and next sentence prediction (NSP) \cite{devlin2018bert}. Normally, standard conditional language models are trained on either left-to-right or right-to-left representations of the context, but MLM used both left-to-right and right-to-left representations of the context. The primary goal of the masked language model is to predict the actual vocabulary identifier of the input's randomly masked tokens. Next sentence prediction (NSP) aids the model in comprehending sentence relationships. This feature helps to improve the performance for the downstream tasks of question-answering (QA) and natural language inference (NLI). To encode the input, the input sentence is first tokenized, and then the tokens are combined with 3 new tokens, namely, CLS, SEP, and MASK. CLS is added at the start of sentence to represent sentence-level classification. To predict the next sentence, SEP is used. During the MLM task, MASK is used to represent masked tokens. English Wikipedia (2,500M words) and the BooksCorpus (800M words) are used for pre-training the BERT model. Summing the corresponding token, segment, and position embeddings yields the input representation for a given token. Primarily, BERT has two variants named as $BERT_{BASE}$ and $BERT_{LARGE}$. $BERT_{BASE}$ has 12 transformer layers, 768 hidden size, 12 attention heads, and 110M total parameters. $BERT_{LARGE}$ has 24 transformer layers, 1024 hidden size, 16 attention heads and 340M total parameters.

\textbf{SciBERT} is a BERT-based pre-trained language model that was trained on a large corpus of scientific text from Semantic Scholar \cite{ammar-etal-2018-construction}. The same size and configuration of $BERT_{BASE}$ is used to train the SciBERT model and allowed 128 tokens of maximum sentence length. SciBERT has 4 variants: \texttt{cased}/\texttt{uncased} and \texttt{basevocab}/\texttt{scivocab}. The \texttt{basevocab} models are fine-tuned from the corresponding $BERT_{BASE}$ models. The \texttt{scivocab} models have been trained from scratch.
\subsection{Pointer-generator network with SciBERT} 
This model consists of a word-embedding layer and a pointer-generator network.  The word-embedding layer converts the words in the input document to embeddings. We have used a pre-trained SciBERT model \cite{beltagy-etal-2019-scibert} to generate word embeddings.
Using this mechanism, each word ($x_t$) in the encoder and the decoder part will be represented as an embedding vector $\vec{x}_t$ as:
\begin{equation} 
   \label{eq_scibert}
   \vec{x}_{t} = g(x_t)
\end{equation}
where $g(.)$ is the embedding-generating function. The CLS token has been added to represent sentence-level classification.
Here, the main use of SciBERT \cite{beltagy-etal-2019-scibert} is that instead of directly feeding the token ids of the input document into the encoder, we are passing the pre-trained SciBERT-generated word embeddings.
In our experiments, the dimension of word embeddings is 768. 
A pointer-generator network \cite{See2017GetTT} augments the sequence-to-sequence (seq2seq) model with attention \cite{nallapati2016abstractive} using a special copying mechanism. When generating words, the decoder probabilistically decides between generating new words from the vocabulary (i.e., from the training corpus) and copying words from the input document (by sampling from the attention distribution). While the generator helps in novel paraphrasing, copying helps to tackle OOV words.
This improves the model's ability to calculate hidden states because the inputs at each time step have been accurately and completely represented, contributing to the improvement of the attention distribution. 
At each decoder time step $t$, the probability of generating a new word is
\begin{equation}\label{eq_pgen}
P_{gen} = \sigma(\vec{W}_{h^{*}}^{\top} \vec{h}_t^{*} + \vec{W}_s^\top \vec{s}_t + \vec{W}_x^\top \vec{x}_t + \vec{b}_{ptr})
\end{equation}
where $\vec{h}_t^*$ is the context vector, $\vec{s}_t$ is the decoder hidden state, $\vec{x}_t$ is the decoder input (which is the decoder output at time $t-1$ during test, and the correct word at time $t-1$ during training), $\sigma$ is the \textit{sigmoid function}, and  $\vec{W}_{h^{*}}$, $\vec{W}_s$, $\vec{W}_{x}$ and $\vec{b}_{ptr}$ are the learnable parameters. 
Hence, for the SciBERT pre-trained embeddings layer the formula in \eqref{eq_pgen} is modified as follows:
\begin{equation}\label{eq_pgen_scibert}
P_{gen} = \sigma(\vec{W}_{h^{*}}^{\top} \vec{h}_t^{*} + \vec{W}_s^\top  \vec{s}_t + \vec{W}_x^\top g(\textbf{x}_t) + \vec{b}_{ptr})
\end{equation}
 
To predict the next word $y_{t}$, the probability distribution over the extended vocabulary (i.e., the fixed vocabulary of the training corpus and the present document) is calculated: 
\begin{equation}
P(y_t) = P_{gen} P_{vocab}(y_t)+(1-P_{gen}) \sum_{i:w_{i}=y_t} a_{t,i}
\end{equation}
where $\vec{a}_t$ is the attention distribution over the fixed vocabulary at time $t$, $a_{t,i}$ is the attention over the  word $w_i$ at time $t$, and $P_{vocab}$ is the probability distribution over the extended vocabulary generated by the softmax layer of the decoder.
The loss for decoder time step $t$ is:
\begin{equation}\label{eq_covfinalloss}
loss_{t} = -\log P(y_{t}^{*}) 
\end{equation}
where $y_t^{*}$ is the  target word. The overall loss for the sequence is the average of the losses over all the decoder time steps for this sequence.

\subsection{Pointer-generator + Coverage mechanism with SciBERT} 
Sometimes the above pointer-generator network redundantly generates the same word multiple times during test. 
The coverage model of Tu et al.  \cite{tu2016modeling} aims to address this problem. This model essentially gives attention to the previous timesteps of the decoder. It computes a coverage vector $\vec{c}^{t}$ defined as the sum of the attention distributions $\vec{a}_{t}$ over all previous timesteps $\tau=1$ to $\tau=t-1$ of the decoder:
\begin{equation}\label{eq_coverageVec}
\vec{c}_{t} = \sum{_{\tau=0}^{^{t-1}}} \vec{a}_{\tau} 
\end{equation}
Note that $\vec{c}_0$ is a zero vector. 
The \textit{coverage vector} will be taken as an extra input to the attention mechanism that is used by the decoder while generating the next word.

The \textit{coverage loss} quantifies if the model is continuously giving more attention to the same words:
\begin{equation}\label{eq_covloss}
CoverageLoss_{t}=\sum_{i} \min({a}_{t,i}, {c}_{t,i})
\end{equation}
Finally, the coverage loss is included in the primary loss function of the decoder. The revised loss for decoder time step $t$ can be written using a hyperparameter $\lambda$ as follows:
\begin{equation}\label{eq_covfinalloss}
loss_{t} = -\log P({y}_{t}^{*}) + {\lambda} \sum_{i} \min(a_{t,i},c_{t,i})
\end{equation}
\section{Experimental Setup}
\label{Experiment}
In this section, we discuss the datasets, the data pre-processing steps, and the experiments.
\subsection{Datasets}

\subsubsection{Dataset of computer science papers}
We use the dataset CSPubSum released by Collins et al. \cite{collins2017supervised} containing URLs of 10142 computer science publications from ScienceDirect\footnote{\url{https://www.sciencedirect.com/}}. 
Every document contains the following fields: title, abstract, research highlights written by the authors, a list of keywords mentioned by the authors, and various sections such as introduction, related work, experiment, and conclusion, as typically found in a research paper.  
We organize each example in this dataset as \textit{(abstract,  author-written research highlights, introduction, conclusion)}. We have observed that, here, the average abstract size is 186 words while that of highlights is 52; and for 98\% of the papers, highlights are 1.5 times or more shorter than the abstract. Thus, at least in terms of the word length, highlights can be considered as a summary not only of the paper but also of the abstract. For our experiments, we split the dataset into train/dev/test in the ratio $80:10:10$, that is, reserve 8115 examples for training, 1014 examples for validation, and 1013 examples for testing. We use this holdout test set to measure the performance of our models on CSPubSum in all cases, except in Section \ref{sec:cv} where we present some results obtained with $K$-fold cross-validation.

\subsubsection{MixSub: a new dataset of papers from multiple domains}
\begin{figure}[!htbp]
   \subfloat[Distribution of train examples.]{\includegraphics[width=.45\textwidth]{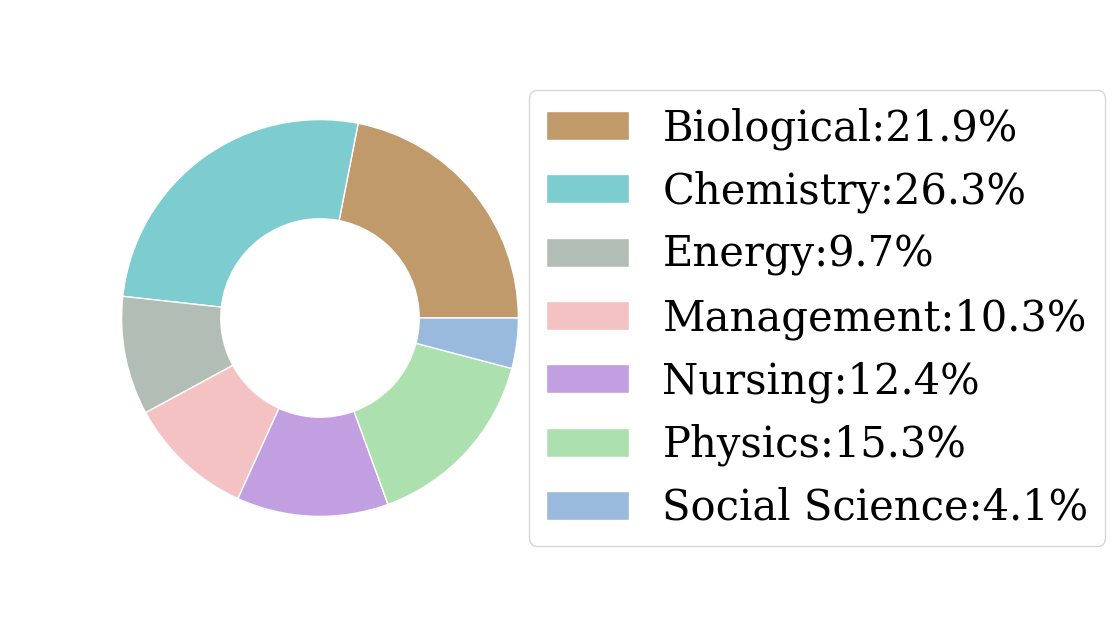}\label{fig:pie_train}}
   \hfill
    \subfloat[Distribution of test examples.]{\includegraphics[width=.45\textwidth]{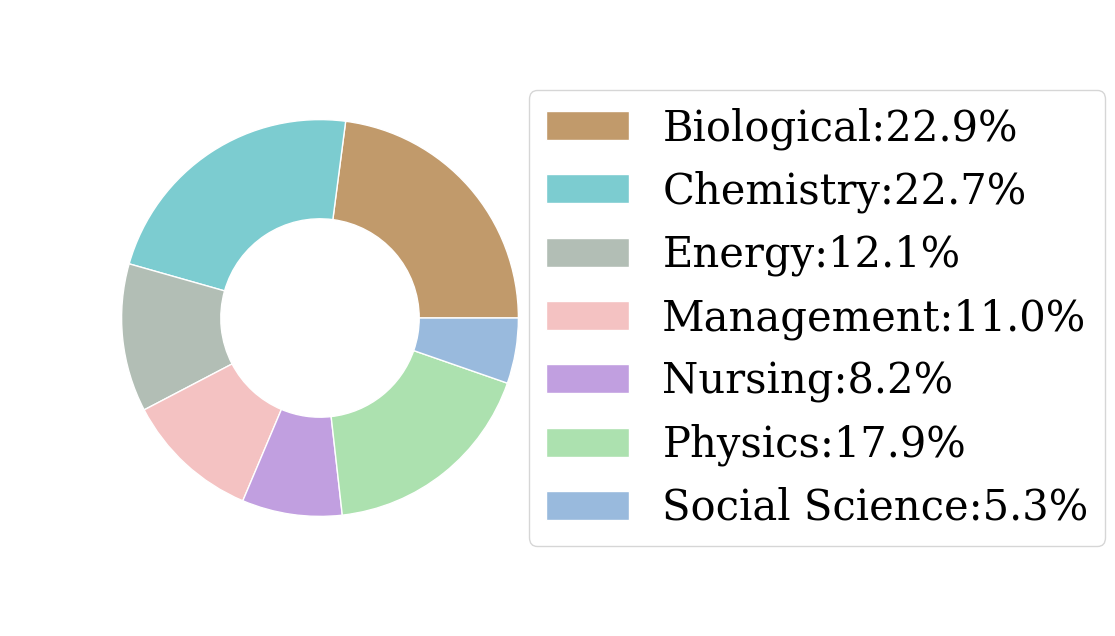}\label{fig:pie_test}}
  \caption{Subject-wise distribution of papers in  MixSub dataset.}\label{fig:pie_MixSub}
\end{figure}

We propose a new dataset called \textbf{MixSub} that contains research articles from multiple domains. Note that the CSPubSum corpus from \cite{collins2017supervised} contains only computer science papers. To prepare MixSub, we crawled the ScienceDirect website and curated articles published in various journals in year 2020. We removed the articles that did not contain research highlights. 
Finally, we got 19785 articles with author-written research highlights as shown in Table \ref{Table:MixSub Dataset}. 
Each example in this dataset is organized as \textit{(abstract,  author-written research highlights)}. We have also segmented the dataset into training, validation and test subsets. 
In this corpus, the average abstract size is 148 words while that of highlights is 57. For 72\% of the papers, highlights are 1.5 times or more shorter than the abstract. We split each category of documents into train/dev/test subsets in the ratio $80:10:10$.
We have grouped similar journal papers according to their domain as shown in Table \ref{Table:MixSub Dataset} and also highlighted using a pie chart Figure \ref{fig:pie_MixSub}.
A summary of the above two datasets is shown in Table \ref{Table:dataset_statistics}.

\begin{table*}[!htbp]
\centering
\caption{Subject-wise URL count in MixSub dataset.}
 \label{Table:MixSub Dataset}
    \begin{adjustbox}{width=.95\linewidth}
{\begin{tabular}{|p{2cm}p{6.8cm}cccc|}  \hline
 Domain Name &Subject name &\#Total & \#Train & \#Val & \#Test \\\hline
\multirow{4}{*}{Biological} &Agricultural and Biological Sciences &2156 &1726 &216 &214\\
&Biochemistry, Genetics and Molecular Biology &976 &806 &71 &99 \\
&Immunology and Microbiology&233 &195 &24 &14\\
&Neuroscience &962 &771 &96 &95 \\\hline

Chemistry&Chemical Engineering &2140 &1713 &214 &213 \\
&Chemistry&2282 &1919 &240 &123\\
&Materials Science &735 &572 &82 &81\\\hline

Energy &Energy &1313 &1025 &145 &143\\
&Environmental Science &677 &517 &81 &79\\\hline

Management &Business, Management and Accounting &698 &560 &70 &68 \\
&Decision Sciences &947 &759 &95 &93\\
&Economics, Econometrics and Finance &421 &324 &56 &41\\\hline

Nursing &Health Sciences &823 &796 &12 &15\\
&Nursing and Health Professions&61 &47 &8 &6\\
&Pharmacology,Toxicology,Pharmaceutical Science &1184 &949 &118 &117\\
&Psychology &28 &24 &3 &1 \\
&Veterinary Science and Veterinary Medicine &186 &156 &19 &11 \\\hline

Physics&Earth and Planetary Sciences &1354 &1038 &159 &157\\
&Mathematics &288 &232 &29 &27\\
&Physics and Astronomy &1469 &1177 &147 &145\\\hline

Social Science&Social Sciences &852 &654 &100 &98\\\hline
\end{tabular} }
\end{adjustbox}
\end{table*}

\begin{table*}[!htbp]
\centering
\caption{Some statistics of CSPubSum and MixSub datasets.}
\begin{adjustbox}{width=.95\linewidth}
{\begin{tabular}{|lcccp{1.56cm}p{1.65cm}p{3cm}|}  \hline
Dataset &Train &Val &Test &Average Words  (abstract) &Average Words (highlight) &\% of article-pairs where compression $>=1.5$ times\\ \hline
CSPubSum &8115 &1014 &1013 &186 &52 &98\\ 
MixSub &15960 &1985 &1840 &148 &57 &72\\\hline 
\end{tabular} }
\label{Table:dataset_statistics}
\end{adjustbox}
\end{table*}
\subsection{Data pre-processing}
We have used the Stanford CoreNLP  Tokenizer\footnote{\url{https://stanfordnlp.github.io/CoreNLP/}} for tokenizing the sentences.
The  whole corpus is first converted to lowercase. 
We have removed all unnecessary symbols, letters, and other elements from the text that do not affect the aim of our research. In particular, HTML tags, parentheses, and special characters have been removed.

Then we reorganized the dataset in several ways to perform various experiments. More specifically, for CSPubSum, we organize it as \textit{(abstract, author-written research highlights)}, \textit{(conclusion, author-written research highlights)}, \textit{(introduction, author-written research highlights)}, \textit{(abstract +  conclusion, author-written research highlights)}, and  \textit{(introduction + conclusion, author-written research highlights)}  where `+' denotes text concatenation. Since the background and a broad summary of the paper normally appear in the introduction, and the main findings of the paper are mentioned in the conclusion, we experiment taking these sections as inputs. Since an overview of the paper is present both in the introduction and the abstract, we do not use them together, rather we use the combinations (abstract +  conclusion), and (introduction + conclusion).
In case of MixSub, we only have \textit{(abstract, author-written research highlights)} because we get the best results on CSPubSum using only abstracts. In future, we might explore the use of full-text or sections from the full-text for MixSub as well.

When the abstract is used as the input, we set the maximum number of input tokens to 400. When the conclusion is used as the input, the maximum number of input tokens allowed is 800. When the introduction is used as the input, the maximum number of input tokens allowed is set to 1200.  For all other inputs, we have restricted the input size to 1500 tokens. In all cases, the maximum token count of the generated research highlights tokens is set to 100. 
The above figures are motivated by the observation that the average length of an abstract is 186, the average length of the author-written highlights in a paper is 52, and the average length of the conclusion is 425, that of the introduction is 837, the average length of  (abstract + conclusion) is 643, and that  of (introduction + conclusion) is 1230. 

\subsection{Implementation details}
We trained four variants of the proposed model: pointer-generator network with word embeddings trained from scratch as part of the model training (\textbf{PGM}), pointer-generator network with coverage mechanism where word embeddings are trained from scratch as part of the model training (\textbf{PGM + Coverage}), pointer-generator network with SciBERT embeddings for the input tokens (\textbf{PGM + SciBERT}), and pointer-generator network with coverage mechanism and SciBERT embeddings for the input tokens (\textbf{PGM + Coverage + SciBERT}). For all variants of SciBERT models, during model training, the embeddings are fine-tuned. 
We trained all models on Tesla P100-PCIE-16GB \texttt{Colab Pro+} that supports GPU-based training. We used mini-batches of size 16. For all models, we used bidirectional LSTMs with cell size of 256. For models without SciBERT, word embeddings of dimension 128 are trained end-to-end with the model. For models with SciBERT, pre-trained word embeddings of dimension 768 are used. For all experiments, we constrained the vocabulary size to the most frequent 50,000 tokens. We considered gradient clipping with a maximum gradient norm of 1.2. Out of the four variations of the SciBERT model, we use \texttt{SciVocab-uncased} \footnote{\url{https://huggingface.co/allenai/scibert_scivocab_uncased/}}. We used other hyperparameters as suggested by \cite{See2017GetTT}. We have used the validation set to determine the number of epochs for training.

\subsection{Evaluation metrics}
To evaluate the performance of the models, we use ROUGH \cite{lin2004rouge}, METEOR \cite{banerjee2005meteor}, and BERTScore \cite{zhang2019bertscore} metrics. These are the standard metrics used to measure the performance of summarization models  \cite{fabbri2021summeval}.
When comparing the model-generated research highlights (\texttt{ModelHighlights}) with the author-written research highlights (\texttt{AuthorHighlights}) for assessment, ROUGE-$n$  calculates the recall, precision, and F1-measure for each model using equations \eqref{Eq:Recall}, \eqref{Eq:Precision} and \eqref{Eq:F1}. Note that an $n$-gram is a contiguous sequence of $n$ words from a piece of text.
Recall ($R$) is defined as:
\begin{equation}\label{Eq:Recall}
    R = \frac{\# matched \ n\-/grams}{\# n\-/grams \ in \ \texttt{AuthorHighlights}}
\end{equation}
Precision ($P$) is defined as:
\begin{equation}\label{Eq:Precision}
    P = \frac{\# matched \ n\-/grams}{\# n\-/grams \ in \ \texttt{ModelHighlights}}
\end{equation}
F1-measure ($F1$) is calculated using the formula:
\begin{equation}\label{Eq:F1}
    F1 = 2*\frac{R*P}{R+P}
\end{equation}
We have used ROUGE-1, ROUGE-2 and ROUGH-L. In particular, ROUGE-L measures the longest matching sub-sequence of words between the two strings. All our ROUGE scores have a 95\% confidence interval of at most $\pm$ 0.25 as reported by the official ROUGE script \cite{lin2004rouge}.

METEOR assigns a score to the match based on a combination of unigram precision, unigram recall, and a fragmentation measure that is intended to directly represent how well-ordered the matched words in the model-generated research highlights and author-written research highlights are. It calculates recall ($R$) and precision ($P$) of unigrams based on  equations \eqref{Eq:Recall} and \eqref{Eq:Precision}, respectively. 
Next, it computes $F_{mean}$ score and chunk penalty using the formula:
\begin{align}
  \label{Eq:Fmean}
    F_{mean} &= \frac{10(R*P)}{R+9P} \\
    Penalty &= 0.5 * \left( \frac{\#chunks}{\#unigrams\_matched} \right)^{3}
\end{align}
where $chunks$ is defined as a set of unigrams that are adjacent in the \texttt{ModelHighlights} and in the \texttt{AuthorHighlights}.
The final METEOR score is computed as follows:
\begin{equation}
  \label{Eq:FinalMeteorScore}
    Score = F_{mean} * (1-Penalty)
\end{equation}
For BERTScore computation, we consider the cosine similarity of contextual embeddings of each word from model-generated research highlights and author-written research highlights, instead of counting the exact words matched across them. Denoting the contextual embeddings of the author-written research highlights  by $\vec{x}=\langle \vec{x}_1, \ldots, \vec{x}_n \rangle$ and those of the model-generated research highlights by $\hat{\vec{x}}=\langle \hat{\vec{x}}_1, \ldots, \hat{\vec{x}}_m \rangle$, the recall ($R_{\BERT}$), precision ($P_{\BERT}$), and F1-scores ($F_{\BERT}$) are calculated as follows:

\begin{equation}\label{Eq:R_{BERT}}
R_{\BERT} = \frac{1}{m} \sum_{\vec{x}_i \in \vec{x}} \max_{ \hat{\vec{x}}_j \in \hat{\vec{x}}} {\vec{x}_i ^\top \hat{\vec{x}}_j}
\end{equation}
\begin{equation}\label{Eq:P_{BERT}}
P_{\BERT} = \frac{1}{n} \sum_{\hat{\vec{x}}_j \in \hat{\vec{x}}} \max_{ \vec{x}_i \in \vec{x}} {\vec{x}_i ^\top \hat{\vec{x}}_j}
\end{equation}

\begin{equation}\label{Eq:F_{BERT}}
    F1_{\BERT} = 2*\frac{R_{\BERT}*P_{\BERT}}{R_{\BERT}+P_{\BERT}}
\end{equation}

\section{Results}
\label{Results}
\subsection{Comparison of pointer-generator type models}
\subsubsection{Evaluation on CSPubSum dataset}

In this sub-section, we report the results of experiments on the CSPubSum dataset for various input types. 

\begin{table*}[!htbp]
\centering
\caption{Evaluation of pointer-generator type models: scores for ROUGE, METEOR and BERTScore on various inputs from CSPubSum dataset. All our ROUGE scores have a 95\% confidence interval of at most $\pm$ 0.25 as reported by the official ROUGE script.} \label{Table:par_all_types_rouge_meteor_bert}
    \begin{adjustbox}{width=.95\linewidth}
{\begin{tabular}{|p{1.8cm}p{4.6cm}ccccc|}  \hline
  & &ROUGE-1 &ROUGE-2 &ROUGE-L &METEOR  &BERTScore\\
Input &Model Name &F1 &F1 &F1 &Final score &F1\\ \hline
\multirow{4}{4em}{abstract only} &PGM &35.44 &11.57 &29.88 &25.4 &83.80\\
&PGM + Coverage &36.57 &12.3 &30.69 &25.4  &84.05\\
&PGM + SciBERT &36.55  &13.44 &33.57 &30.34 &86.34\\ 
&PGM + Coverage + SciBERT &\bf{38.26} &\bf{14.26} &\bf{35.51}  &\bf{32.62} &\bf{86.65}\\\hline
\multirow{4}{4em}{conclusion only} &PGM &32.11 &9.32 &29.62 &24.04 &85.72\\
&PGM + Coverage &34.33  &9.73 &31.71 &24.99 &86.07\\
&PGM + SciBERT &33.19 &9.8 &30.49 &24.26 &86.03\\ 
&PGM + Coverage + SciBERT  &\bf{34.81} &\bf{10.02} &\bf{32.31} &\bf{25.21} &\bf{86.52}\\\hline
\multirow{4}{4em}{introduction only} &PGM &30.85 &7.92 &28.55 &19.76 &85.25\\
 &PGM + Coverage &32.46 &8.18 &30 &20.50 &85.48\\
&PGM + SciBERT &31.56 &8.79 &29.18 &23.09 &85.93\\ 
&PGM + Coverage + SciBERT &\bf{33.33} &\bf{9.7} &\bf{30.86} &\bf{24.10} &\bf{86.17}\\\hline
\multirow{4}{4em}{abstract + conclusion} &PGM &29.85 &8.16 &25.80 &19.38 &83.19\\
 &PGM + Coverage  &31.70 &8.31 &26.72 &20.92 &83.49\\
 &PGM + SciBERT  &32.84 &9.86 &30.34 &24.59 &86.13\\
 &PGM + Coverage + SciBERT &\bf{35.09} &\bf{10.94} &\bf{32.69} &\bf{27.31} &\bf{86.52}\\\hline
\multirow{4}{4em}{introduction + conclusion} &PGM &29.78 &7.47 &25.15 &19.25 &83.05\\
 &PGM + Coverage &31.63 &7.65 &26.25 &20.24 &83.32\\
&PGM + SciBERT &32.38 &9.63 &29.79 &23.95 &86.11\\
&PGM + Coverage + SciBERT &\bf{35.32} &\bf{10.93} &\bf{32.76} &\bf{26.57} &\bf{86.59}\\\hline
\end{tabular} }
\end{adjustbox}
\end{table*}

\noindent \textbf{Input: Abstract:} 
Results are shown in Table \ref{Table:par_all_types_rouge_meteor_bert} for ROUGE-1, ROUGE-2, ROUGE-L, METEOR and BERTScore when the input is the abstract of a research paper.
We observe that among the four models, the pointer-generator network with coverage mechanism and SciBERT (\textbf{PGM + Coverage + SciBERT}) achieve the highest ROUGE, METEOR and BERTScore values.\\
\noindent \textbf{Input: Conclusion:} 
Results are shown in Table \ref{Table:par_all_types_rouge_meteor_bert} for ROUGE-1, ROUGE-2, ROUGE-L, METEOR  and BERTScore when the input is only the conclusion of a research paper.
We observe that among the four models, the (PGM + Coverage + SciBERT) model achieves the highest ROUGE, METEOR and BERTScore values.\\
\noindent \textbf{Input: Introduction:} 
Results are shown in Table \ref{Table:par_all_types_rouge_meteor_bert} for ROUGE-1, ROUGE-2, ROUGE-L, METEOR  and BERTScore when the input is  the introduction of a research paper.
We observe that among the four models, the (PGM + Coverage + SciBERT) model achieves the highest ROUGE, METEOR  and BERTScore values.\\
\noindent \textbf{Input: Abstract + Conclusion:} 
Results are shown in Table \ref{Table:par_all_types_rouge_meteor_bert} for ROUGE-1, ROUGE-2, ROUGE-L scores, METEOR  and BERTScore when the input is the combination of the abstract and the conclusion of a paper. 
We again observe that the best performance is achieved by the (PGM + coverage + SciBERT) model.\\
\noindent \textbf{Input: Introduction + Conclusion:} 
When the inputs is the combination of introduction and conclusion in the test dataset, we record ROUGE-1, ROUGE-2, ROUGE-L scores, METEOR and BERTScore  as shown in Table \ref{Table:par_all_types_rouge_meteor_bert}. 
The best performing model is (PGM + coverage + SciBERT).
Upon analysis of the dataset, we found that in many cases the highlights are largely included in the `abstract'; therefore, using the `abstract' as input to the model results in high performance. 
We have observed that the `conclusion' typically presents a more detailed and technically dense  description of the findings in contrast to the more overview-style summary included in the research highlights (see, for example,  these papers\footnote{\url{https://www.sciencedirect.com/science/article/abs/pii/S0010448514001870}} \footnote{\url{https://www.sciencedirect.com/science/article/pii/S0010448514001638}}). The `conclusion' also includes future work, which does not form part of the highlights. So adding the `conclusion' with the `abstract' does not improve the performance.  Although the `introduction' of a paper often contains the main findings of the paper, it also contains a lot of other information (typically, to build the background and context to the current work) that is not included in the highlights and must be filtered away by the model when generating the output. 

\subsubsection{$K$-Fold cross-validation}
\label{sec:cv}
\begin{table*}[!htbp]
\centering

\caption{$K$-fold cross-validation of the proposed models on CSPubSum dataset. For comparison, the performance of the models with holdout validation are reproduced from Table \ref{Table:par_all_types_rouge_meteor_bert}.} 
\label{Table:K-Fold_par_all_types_rouge_meteor_bert}
    \begin{adjustbox}{width=.95\linewidth}
{\begin{tabular}{|p{1.8cm}p{4.6cm}ccccc|}  \hline
  & &ROUGE-1 &ROUGE-2 &ROUGE-L &METEOR  &BERTScore\\
Input &Model Name &F1 &F1 &F1 &Final score &F1\\ \hline
\multirow{4}{8em}{abstract only (holdout validation)} &PGM &35.44 &11.57 &29.88 &25.4 &83.80\\
&PGM + Coverage &36.57 &12.3 &30.69 &25.4  &84.05\\
&PGM + SciBERT &36.55  &13.44 &33.57 &30.34 &86.34\\ 
&PGM + Coverage + SciBERT &\bf{38.26} &\bf{14.26} &\bf{35.51}  &\bf{32.62} &\bf{86.65}\\\hline

\multirow{2}{8em}{abstract only (5-fold CV)} &PGM + SciBERT &37.79 &12.77 &34.78 &29.92 &86.72\\ 
&PGM + Coverage + SciBERT&\bf{39.43} &\bf{15.25} &\bf{36.48}  &\bf{30.85} &\bf{87.01}\\\hline

\end{tabular} }
\end{adjustbox}
\end{table*} 

We also perform $K$-fold cross-validation (CV)  of our model (\textbf{PGM + Coverage + SciBERT}) on the CSPubSum dataset. For this purpose, we set $K=5$, that is, we split the whole dataset into five distinct parts. We trained using four parts (or folds) and tested the model using the remaining part. In each case, we trained the pointer-generator network with SciBERT for 20000 iterations, then added the coverage mechanism and continued training for another 1000 iterations. In all cases, we consider only the abstracts of the  CSPubSum dataset as the input. Table \ref{Table:K-Fold_par_all_types_rouge_meteor_bert} reports the ROUGE, METEOR and BERTScore for the model (PGM + Coverage + SciBERT) with 5-fold cross-validation and compares the performance with that of holdout validation. Since $K$-fold cross-validation is computationally quite expensive, we did not conduct it for the other input types. Note that the performance achieved by $K$-fold cross-validation is slightly higher than that reported by holdout validation. Since it is widely believed (see, for example, \cite{10.5555/1643031.1643047,raschka2018model})
 that $K$-fold cross-validation results are a better indicator of the generalization performance, our model is likely to be better than that indicated by holdout testing.   
\subsubsection{Comparison with previous works}
\begin{table*}[!htbp]
\centering
\caption{Comparison of the performance of the proposed model with that of other approaches for CSPubSum dataset. \label{Table:per_comparison_others}}
\begin{adjustbox}{width=.95\linewidth}
{\begin{tabular}{|lccccc|}  \hline
Model Name &ROUGE-1 &ROUGE-2 &ROUGE-L &METEOR &BERTScore\\
 &F1 &F1 &F1 &Final score &F1\\ \hline
LSTM Classification \cite{collins2017supervised} &\textemdash &12.7 &29.50 &\textemdash &\textemdash\\ 
Gradient Boosting Regressor \cite{cagliero2020extracting} &\textemdash &13.9 &31.60 &\textemdash &\textemdash\\ 
PGM + Coverage + GloVe \cite{rehman2021automatic} &31.46 &8.57  &29.14 &12.01 &85.31\\
NER + PGM + Coverage \cite{rehman-etal-2022-named} &38.13 &13.68 &35.11  &31.03 &86.3\\
PGM + Coverage + SciBERT (ours) &\bf{38.26} &\bf{14.26}  &\bf{35.51} &\bf{32.62} &\bf{86.65}\\\hline
\end{tabular}}
\end{adjustbox}
\end{table*}

Table \ref{Table:per_comparison_others} compares the performance of our proposed approach (\textbf{PGM + Coverage + SciBERT}) with other competitive baselines in the literature, namely, an \textbf{LSTM-based} extractive summarization model  \cite{collins2017supervised}, a \textbf{gradient boosting regression} extractive summarization model \cite{cagliero2020extracting}, and a \textbf{PGM model with GloVe embeddings} for abstractive summarization \cite{rehman2021automatic}, on the CSPubSum dataset in terms of the ROUGE-1 (F1), ROUGE-2 (F1), ROUGE-L (F1), METEOR and BERTScore (F1) metrics. 

ROUGE-2 F1-score and ROUGE-L F1-score of the LSTM-based model in \cite{collins2017supervised} are 12.7 and 29.50, respectively while those in the gradient boosting regression model \cite{cagliero2020extracting} are 13.9 and 31.6, respectively. Both the above methods use extractive summarization on the full text (sans abstract) of the paper, that is, they select a set of sentences from a given document for inclusion in the research highlights. Rehman et al. \cite{rehman2021automatic} use abstractive summarization to generate research highlights from abstracts only, and the best performing model in it is a pointer-generator network with coverage and GloVe embeddings that records ROUGE-1 F1, ROUGE-2 F1, ROUGE-L F1, METEOR score, and BERTScore F1 values as 31.46, 8.57, 29.14, 12.01 and 85.31, respectively. In a follow-up work, Rehman et al. \cite{rehman-etal-2022-named} combined named-entity recognition (NER) with coverage-augmented pointer-generator network to generate research highlights from different parts of a paper. The best performing model in \cite{rehman-etal-2022-named} is denoted as (NER + PGM + Coverage) in Table \ref{Table:per_comparison_others}, and it uses only the abstract; it produces ROUGE-1 F1, ROUGE-2 F1, ROUGE-L F1, METEOR, and BERTScore F1 values as 38.13, 13.68, 35.11, 31.03 and 86.3, respectively. 
We clearly observe that the method proposed in this paper, i.e., a pointer-generator network with coverage and SciBERT word embeddings, achieves the best ROUGE-2 F1-score and ROUGE-L F1-score which are 14.26 and 35.51, respectively. The same model also achieves the highest METEOR score and BERTScore F1 of 32.62 and 86.65, respectively, among the three pointer-generator models compared in Table \ref{Table:per_comparison_others}. (Note that here we have measured the performance on the holdout test set.) The above model (PGM + Coverage + SciBERT) uses only the abstracts as input unlike the methods in \cite{collins2017supervised} and \cite{cagliero2020extracting}, that use the full text of the paper. Abstracts being much shorter than the main text of a paper, the computational overhead is  significantly reduced. Our method establishes a new state-of-the-art for the CSPubSum dataset.

\subsubsection{Evaluation on MixSub dataset}
In this sub-section, we report the results of experiments on the MixSub dataset. 
We trained the models in two ways: 
\begin{itemize}
    \item \textbf{Case 1:} We trained all the four models on each subject cluster separately and tested them on the corresponding test documents.
    \item \textbf{Case 2:} We did not distinguish between the subject categories of the papers but simply collected all the documents of the training corpus, and trained the models. Then we evaluated them on the test corpus and reported the results for each subject category.
\end{itemize}
Note that in each case, the input is only the abstract of a paper. Since MixSub currently does not contain the body of a paper, we cannot use other sections of a paper as the input. 
Results are reported in Table \ref{Table:par_clusterMixSub_rouge_meteor_bert} for ROUGE-1, ROUGE-2, ROUGE-L, METEOR and BERTScore. The top row labeled `Full MixSub' shows the results when the models are trained on the whole training corpus without regard to the specific subject category of the papers and tested on the test corpus, again without regard to the specific subject category of the papers. The remaining rows show the scores obtained on each category of papers when the models are trained either on the respective clusters (Case 1) or on the whole training corpus without regard to subject category (Case 2).   
We observe that among the four models, (PGM + coverage + SciBERT) achieves the highest ROUGE, METEOR and BERTScore values. We observe that sometimes training on subject-specific clusters leads to higher scores and at other times, training on the whole corpus produces better scores at the subject level. But (\textbf{PGM + Coverage + SciBERT}) outperforms all the other models in all cases.
\begin{table*}[!htbp]
\centering 
\caption{Evaluation of pointer-generator type models: scores for ROUGE, METEOR and BERTScore on  MixSub dataset. The first row (where dataset is `Full MixSub') indicates the performance when the models are trained on the whole MixSub training set and evaluated on the whole MixSub test set, without distinguishing between the subject categories of the papers. In the remaining part of the table, two cases are considered: 
Case 1: Trained on each subject-cluster of MixSub training set and evaluated on the corresponding test set;  Case 2: Trained on the entire MixSub training set and evaluated on each subject-cluster of MixSub test set.}
\label{Table:par_clusterMixSub_rouge_meteor_bert}
    \begin{adjustbox}{width=.95\linewidth}
   
{\begin{tabular}{|p{1.8cm}p{1cm}p{4.6cm}ccccc|}  \hline
 & & &ROUGE-1 &ROUGE-2 &ROUGE-L &METEOR  &BERTScore\\
Dataset &Input &Model Name &F1 &F1 &F1 &Final score &F1\\ \hline
\multirow{4}{*}{Full MixSub} &\multirow{4}{4em}{abstract only} &PGM  &29.3 &8.43 &26.99 &21.46 &83.41\\
& &PGM + Coverage &31.52 &9.18 &29.21 &22.91 &85.22\\
& &PGM + SciBERT &30.44 &9.68 &27.81 &23.38 &84.83\\
& &PGM + Coverage + SciBERT &\bf{31.78} &\bf{9.76} &\bf{29.3} &\bf{24} &\bf{85.25}\\\hline
\multirow{8}{*}{Biological} &Case 1 &PGM &25.4 &5.13  &23.56  &18.02 &83.51\\
 &Case 2 &PGM &27.88 &7.36 &25.77 &9.29 &81.01\\
&Case 1 &PGM + Coverage &28.23 &6.18 &25.96 &19.60 &83.99\\
&Case 2 &PGM + Coverage &28.76 &7.74	&26.76 &9.89 &82.01\\
&Case 1 &PGM + SciBERT &28.74 &7.45 &26.56 &20.87  &84.49\\
&Case 2 &PGM + SciBERT &28.42	&8.01 &26.02 &9.92 &81.1\\
&Case 1 &PGM + Coverage + SciBERT &\bf{29.9}  &7.6  &\bf{27.57}  &\bf{21.53} &\bf{84.74}\\
&Case 2 &PGM + Coverage + SciBERT &28.88 &\bf{8.03} &26.76 &9.98 &84.72\\\hline
\multirow{8}{*}{Chemistry} &Case 1 &PGM &27.83 &7.39 &26.1 &13.58 &82.99\\
&Case 2 &PGM &27.44 &7.15 &25.55 &9.63 &81.83\\
&Case 1 &PGM + Coverage &29.67  &8.09  &27.4 &14.14 &83.02\\
&Case 2 &PGM + Coverage &29.68 &7.73 &27.57 &9.88 &82.11\\
&Case 1 &PGM + SciBERT  &30  &8.58  &27.98 &22.38 &84.93 \\
&Case 2 &PGM + SciBERT  &29.16 &8.47 &26.96 &9.85 &81.92 \\
&Case 1 &PGM + Coverage + SciBERT &\bf{31.4} &\bf{8.9}  &\bf{29.33} &\bf{24.37} &\bf{85.11} \\
&Case 2 &PGM + Coverage + SciBERT &30.41 &8.58 &28.15 &10.19 &82.19 \\\hline
\multirow{8}{*}{Energy} &Case 1 &PGM &23.81 &4.21 &21.99 &15.80 &83.44\\
&Case 2 &PGM &29.33 &8.56 &26.91 &9.40 &81.07\\
&Case 1 &PGM + Coverage  &27.12 &4.71 &24.98 &18.11 &84.05\\
&Case 2 &PGM + Coverage  &31.69 &9.24 &29.25 &10.42 &82.55\\
&Case 1 &PGM + SciBERT &28.87 &6.18 &26.5 &19.92 &84.53\\
&Case 2 &PGM + SciBERT &29.61 &9.09 &26.62 &9.70 &81.9\\
&Case 1 &PGM + Coverage + SciBERT &30.04 &6.84 &27.4 &\bf{20.85} &\bf{85.51}\\
&Case 2 &PGM + Coverage + SciBERT &\bf{32.15} &\bf{9.66} &\bf{29.7} &10.77 &82.84\\\hline
\multirow{8}{*}{Management} &Case 1 &PGM  &32.39 &8.73  &30.08 &15.80 &83.18\\
&Case 2 &PGM &34.51 &11.68 &31.64 &9.89 &81.53\\
&Case 1 &PGM + Coverage  &34.47 &9.54  &31.77 &18.23 &83.67 \\
&Case 2 &PGM + Coverage  &37.25 &13.23 &34.4 &10.54 &82.19 \\
&Case 1 &PGM + SciBERT  &33.54 &9.15 &30.9 &23.17 &85.20 \\
&Case 2 &PGM + SciBERT  &35.65 &13.18 &32.81 &10.66 &82.26 \\
&Case 1 &PGM + Coverage + SciBERT &36.05 &11.02 &33.27 &\bf{25.24}  &\bf{85.66} \\
&Case 2 &PGM + Coverage + SciBERT &\bf{38.39} &\bf{13.62} &\bf{35.64} &11.39 &83.03\\\hline
\multirow{8}{*}{Nursing} &Case 1 &PGM &25.2 &4.82 &22.95 &17.01 &83.18\\
&Case 2 &PGM &28.64 &7.9 &26.08 &8.55 &81.08\\
&Case 1 &PGM + Coverage &28.2 &5.46 &25.5 &18.79 &83.71\\
&Case 2 &PGM + Coverage &30.38 &8.39 &27.83 &9.78 &81.61\\
&Case 1 &PGM + SciBERT &30.21 &6.71 &27.54 &21.16 &84.21\\
&Case 2 &PGM + SciBERT &31.43 &9.43 &28.42 &9.52 &81.04\\
&Case 1 &PGM + Coverage + SciBERT &\bf{31.61}  &8.09 &28.7 &\bf{22.73} &\bf{84.57}\\
&Case 2 &PGM + Coverage + SciBERT &\bf{31.61} &\bf{10.28} &\bf{28.83} &9.98 &81.42\\\hline
\multirow{8}{*}{Physics} &Case 1 &PGM &29.97 &7.98 &27.44 &21.19 &84.19\\
&Case 2 &PGM &30.41 &9.07 &28.07 &10.21 &81.05\\
&Case 1 &PGM + Coverage &31.06 &7.9 &28.52 &21.26 &84.95\\
&Case 2 &PGM + Coverage &32.05 &10.26 &30.27 &10.51 &82.4\\
&Case 1 &PGM + SciBERT &30.67 &8.3 &28.45  &21.76 &85.02\\
&Case 2 &PGM + SciBERT &31.31 &10.67 &28.6 &10.38 &81.93\\
&Case 1 &PGM + Coverage + SciBERT &32.13 &8.92 &29.53 &\bf{22.83} &\bf{85.25}\\
&Case 2 &PGM + Coverage + SciBERT &\bf{32.99} &\bf{11.01} &\bf{30.35} &11.16 &82.45\\\hline
\multirow{8}{*}{Social Science} &Case 1 &PGM &22.64 &4.36 &20.51 &13.94 &81.86\\
&Case 2 &PGM &30.23 &10.39 &27.29 &9.23 &81.64\\
&Case 1 &PGM + Coverage  &26.96 &5.17 &24.19 &16.08 &82.72\\
&Case 2 &PGM + Coverage  &31.99 &11.2 &28.82 &9.61 &81.69\\
&Case 1 &PGM + SciBERT &30.11 &8 &26.87 &19.02 &83.35\\
&Case 2 &PGM + SciBERT &31.21 &10.97 &27.75 &9.63 &81.65\\
&Case 1 &PGM + Coverage + SciBERT &31.89 &8.67 &28.36 &\bf{19.99} &\bf{83.81}\\
&Case 2 &PGM + Coverage + SciBERT &\bf{32.35} &\bf{11.8} &\bf{28.87} &9.79 &81.75\\\hline
\end{tabular} }
\end{adjustbox}
\end{table*}

\subsection{Comparison with pre-trained models}
We have chosen the following pre-trained models from the Hugging Face website for the purpose of comparison: \textbf{T5-base}\footnote{\url{https://huggingface.co/t5-base}},  \textbf{Distilbart-CNN-12-6}\footnote{\url{https://huggingface.co/sshleifer/distilbart-cnn-12-6}}, \textbf{GPT-2}\footnote{\url{https://huggingface.co/gpt2}} and \textbf{ProphetNet-large-uncased-cnndm}\footnote{\url{https://huggingface.co/microsoft/prophetnet-large-uncased-cnndm}}. 
We fine-tuned all four models to 15 epochs with CSPubSum where 8115 documents (each comprising an abstract and author written research highlights) are taken for training. 
We tested them on the test dataset of 1013 examples. We used a batch size of 4 for fine-tuning all four pre-trained models.
Observations on the test set are shown in Table \ref{Table:per_finetuning_test}. 
The performance of ProphetNet-large-uncased-cnndm pretrained model is significantly worse than that of other models; the training duration and compute resources we used appeared to be inadequate for this model. 
We observe that T5-base performs better than the other models in terms of ROUGE and BERTScore metrics while Distilbart-CNN-12-6 gives the highest METEOR score.  The slight performance gain of pre-trained models is not surprising at all given the number of parameters and the exhaustiveness of the training of such models. Rather  the closeness of the proposed model, which does not require fine-tuning a large pre-trained transformer model, appears to demand more attention to strike the right trade-off between performance and the resources needed for training. 

In the next sub-section, we will discuss an important aspect of these large models, which has received attention in the recent years. This aspect deals with the energy efficiency of algorithms that is also related to the consequent carbon footprint.\\


\begin{table*}[!htbp]
\centering 
\caption{Performance of fine-tuned versions of pre-trained models on CSPubSum dataset using abstracts of the papers as the input. The highest performance scores are marked with bold.
\label{Table:per_finetuning_test}}
\begin{adjustbox}{width=.95\linewidth}
{\begin{tabular}{|lccccc|}  \hline
{Model Name} &ROUGE-1 &ROUGE-2 &ROUGE-L &METEOR &BERTScore\\
 &F1 &F1 &F1 &Final score &F1\\ \hline
T5-base & {\bf 40.03} & {\bf 16.27} & {\bf 37.64} &36.33 & {\bf 86.80} \\
Distilbart-CNN-12-6 &39.95 &16.13 &37.16 & {\bf 38.99} &86.69  \\
GPT-2 &33.12 &11.76 &30.64 &33.14 &85.30  \\
ProphetNet-large-uncased-cnndm &23.95 &0.96 &20.38 &15.3 &81.41  \\
PGM + Coverage + SciBERT (ours) &38.26  &14.26 &35.51 &32.62 &86.65\\ \hline
\end{tabular} }
\end{adjustbox}
\end{table*}

\subsection{Analysis of energy consumption}
Recently transformer architectures have significantly improved  the performance of various natural language processing (NLP) tasks. Inspired by the original transformer \cite{vaswani2017attention},
language models such as ELMo \cite{DBLP:journals/corr/abs-1802-05365}, BERT \cite{devlin2018bert}, GPT family \cite{radford2018improving} and BART \cite{lewis-etal-2020-bart} have emerged and produced state-of-the-art performance on various tasks. However, they require enormous amounts of data and compute resources for pre-training. This large computation consumes a lot of energy and has a high carbon footprint. It has an adverse financial and environmental impact \cite{strubell2019energy, lannelongue2021green}.

The expression to calculate carbon footprint $C$ (in gram carbon dioxide equivalent or gCO$_2$e) as given in the equation \ref{eq:eq_carbon_footprint} is taken from \cite{lannelongue2021green}.

\begin{equation}\label{eq:eq_carbon_footprint}
C=t\times(n_c\times P_c \times u_c + n_m \times P_m) \times PUE \times CI \times 0.001 
\end{equation}

We modified Equation \ref{eq:eq_carbon_footprint} to Equation \ref{eq:Our_carbon_footprint}:
\begin{align}\label{eq:Our_carbon_footprint}
C &=  t \times (n_c\times P_c \times u_c + n_{gpu} 
   \times P_{gpu} \times u_{gpu}  \nonumber \\
   & + n_m \times P_m) 
    \times PUE \times CI \times 0.001 
\end{align}
where $t$ is the running time (in hours), $n_c$ is the number of cores, $P_c$ is the power draw of a computing core, $u_c$ is the core usage factor (between 0 to 1), $n_{gpu}$ is the number of GPUs, $P_{gpu}$ is the power drawn by the GPU, $u_{gpu}$ is the GPU usage  factor (between 0 to 1), $P_m$ is the power draw of a memory unit (in watt). The power draw of memory is considered as 0.3725 W per GB \cite{karyakin2017analysis,lannelongue2021green}. 

We trained all the models on Tesla P100-PCIE \texttt{Colab Pro+} that supports GPU. 
The  efficiency coefficient of the data center is known as PUE (power usage effectiveness). Google uses ML to reduce its global yearly average PUE to 1.10 \cite{PUE}. We use average worldwide value as carbon intensity (CI) of 475 gCO$_2$e KW/hour\cite{CarbonIntensity}.
Gross CO$_2$ emission during training for T5 pre-trained model \cite{10.5555/3455716.3455856} was 46.7 tCO$_2e$ \cite{patterson2021carbon}, any transformer$_\text{big}$ model training required 192 lbsCO$_2e$ \cite{strubell2019energy} and BERT base model with GPU required 1438 lbsCO$_2$e \cite{strubell2019energy}. 
We measure memory and compute power consumption and emission of CO$_2$ footprint using the WandB tool \footnote{\url{https://wandb.ai/site}}. The quantitative results are shown in Table \ref{Table:power_consumption_co2emission}. In our proposed model, we require SciBERT embeddings of the input documents as input. So as a pre-processing step before model training, we encode the documents with SciBERT: this is a one-time operation and not repeated in every epoch. 
Table \ref{Table:power_consumption_co2emission} clearly shows that our proposed model (third column) has fewer trainable parameters, and lower computational overhead and smaller carbon footprint per epoch than those of the other models.  
We have graphically compared the \% of GPU utilization, \% of CPU utilization, GPU Power usage, GPU memory allocated, memory used by process and required process CPU threads of the models over the training duration in Figure  \ref{fig:resource_utilization}. 
The figure shows that GPU and CPU utilization, GPU power usage, and the process memory used by our proposed model are lower than those used in fine-tuning the large pre-trained summarization models. While our model consumes a large memory for a short time, the other models typically have a larger memory consumption that remains steady for a longer duration. Our model exploits more CPU threads than GPT-2 but fewer threads than other compared models. We believe that researchers should give attention to energy-friendly models and algorithms rather than only to performance metrics. In this context, our model is a better alternative to large pre-trained transformers.

\begin{table*}[!htbp]
\centering
\caption{Power consumption, compute expenditure, and CO$_2$ emission statistics for summarization models. \label{Table:power_consumption_co2emission}}
\begin{adjustbox}{width=.95\linewidth}
{\begin{tabular}{|p{1.9cm}cp{1.6cm}p{1.3cm}p{1.2cm}p{1.4cm}p{1.3cm}|}  \hline
{Factors} &{Sub-Factor} & PGM + Coverage + SciBERT 
 & ProphetNet-large-uncased-cnndm & GPT-2 &  Distilbart-CNN-12-6  & T5-base\\\hline
\multicolumn{2}{|c}{Total trainable  parameters}  &21.5M  &391M &117M &305M &220M  \\\hline
\multirow{3}{*}{Colab Notebook} & Avail. RAM: 51GB &2.86GB &4.07GB &3.20GB &6.47GB &3.89GB \\
&Avail. GPU: 16GB &1.14GB &15.57GB &13.51GB &14.64GB &2.92GB\\
&Avail. Disk: 166.83GB &48.30GB &41.53GB &40.64GB &41.21GB &44.57GB\\\hline
\multirow{2}{*}{Power consumed} & Max. GPU power: 250W &116W  &189W &187W &172W &206W \\
& Max. CPU power: 95W &95W &95W &95W &95W &95W \\\hline
\multicolumn{2}{|c}{\% of GPU utilization}  &75\% &100\% &97\% &100\% &97\%  \\\hline
\multicolumn{2}{|c}{\% of GPU memory allocated}  &55\% &99\% &89\% &92\% &60\%  \\\hline
\multicolumn{2}{|c}{\% of CPU utilization}  &35\%   &16\%  &27\%  &19\%  &17\%   \\\hline
\multicolumn{2}{|c}{Used process CPU threads}  &54 &57 &38 &55 &66  \\\hline
\multicolumn{2}{|c}{Process memory in use (GB)}  &0.834   &22.28  &3.84  &19.18  &12.42   \\\hline
\multicolumn{2}{|c}{Time for one epoch (mins)}  &5.17 & 31 & 22 & 19 & 15  \\\hline
\multicolumn{2}{|c}{Epoch-wise carbon footprint (gms/epoch)}  &5.56 &56.72 &40.68 &32.35 &28.93  \\\hline
\end{tabular}}
\end{adjustbox}
\end{table*}

\Figure[t!](topskip=0pt, botskip=0pt, midskip=0pt)[width=.85\textwidth]{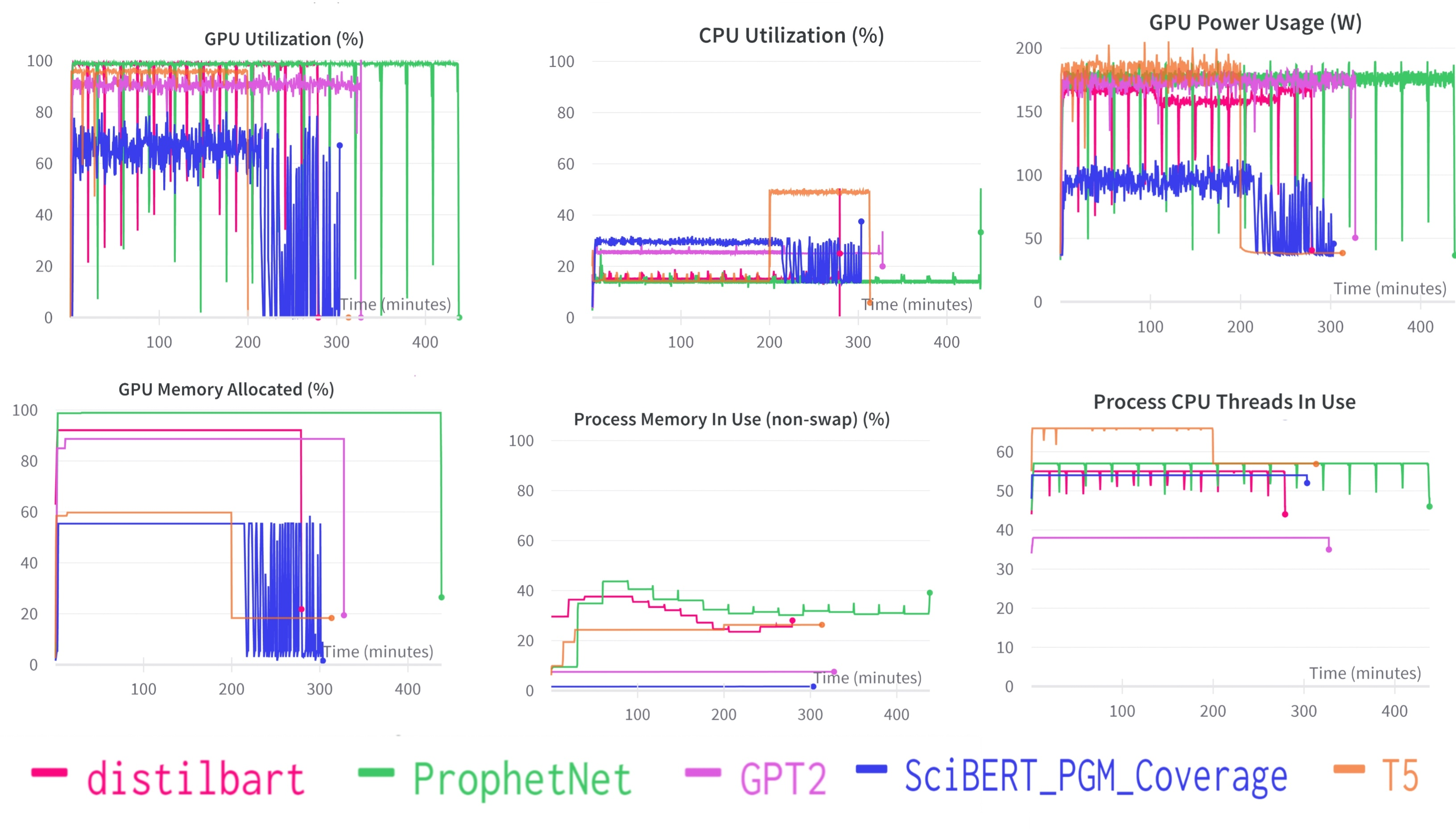}
{Comparison of compute resources used by summarization models.\label{fig:resource_utilization}}

\section{Case Studies}
\label{Case Studies}
\subsection{Case study on CSPubSum dataset}
We now present a few examples demonstrating the outputs produced by the pointer-generator type models used in this paper.
In all the case studies reported below, \textit{yellow} color represents \colorbox{backgG}{factual errors}
and \textit{orange} shows \colorbox{backgPm}{repeating words}. Figure \ref{fig:sample_Abs_RHS} illustrates the highlights produced by the four models when the input is only the abstract.  Note that the vanilla pointer-generator network misses or incorrectly uses some keywords while generating the highlights. For example, it produces ``... algorithm for expression data clustering.'' instead of ``... algorithm for microarray gene expression data clustering.'', and ``... type 2 fuzzy means'' instead of ``... type 2 fuzzy c-means''. While use of SciBERT corrects these issues, unnecessarily repeated words are seen when coverage mechanism is absent. The output produced by the (\textbf{PGM + Coverage + SciBERT}) model is closest to the author-written highlights. 

Figure \ref{fig:sample_Abs+Con_RHS} and Figure \ref{fig:sample_Int+Con_RHS} depict the highlights produced by the models when the input is (abstract + conclusion) and (introduction + conclusion), respectively. We observe that the highlights produced by all the models for the last four input types contain a number of acronyms like `fcm' (fuzzy C-means), `gt2' (general type 2), `fss' (fuzzy sets), and `cvi' (cluster validity index) which occur frequently in the introduction and conclusion of the paper. Since the abstract typically does not contain acronyms, highlights generated using it are also generally free of acronyms.
\begin{figure*}[!tbp] 
\centering 
\begin{tabular}{ |p{17cm}|} \hline
{\bf Author-written research highlights:} Presenting a new two stage meta heuristic clustering algorithm based on general type 2 fuzzy sets. Incorporating a new similarity based objective function using alpha plane representation of general type 2 fuzzy sets. Implementing the proposed approach on real microarray gene expression datasets.\\\hline
{\bf PGM:} We propose a new two stage meta heuristic algorithm for \colorbox{backgG}{expression data} clustering. New objective function for general type 2 \colorbox{backgG}{fuzzy means} is proposed. Output is inserted to the first stage until no better solution is obtained. Proposed approach is evaluated using three microarray gene expression datasets. Experiments verify the performance of the proposed approach.\\\hline 	    
{\bf PGM + Coverage:} A novel meta heuristic methodology for microarray gene expression data clustering is proposed. Two stage inspired the \colorbox{backgPm}{two stage} meta heuristic optimization framework simulated annealing is proposed. Philosophy of the meta heuristic framework is devoted to the annealing process accompanied by its proposed perturbation mechanisms. Experiments were conducted using three microarray gene expression datasets. Experiments demonstrate the capabilities of the proposed approach.\\ \hline
{\bf PGM + SciBERT:} A new two stage meta heuristic algorithm is proposed. A new objective function using planes for general type 2 fuzzy sets is presented. \colorbox{backgPm}{A new objective function using planes for general type 2 fuzzy} C means clustering algorithm is proposed. The proposed approach has been evaluated using several synthesized datasets and three microarray gene expression datasets. \colorbox{backgPm}{The proposed approach is evaluated using several synthesized datasets} \colorbox{backgPm}{and three microarray gene expression datasets}.\\\hline
{\bf PGM + Coverage + SciBERT:} A powerful data clustering approach based on the concept of planes in general type 2 fuzzy sets is presented. A new objective function using planes for general type 2 fuzzy c means clustering algorithm is proposed. The proposed approach has been evaluated using three microarray gene expression datasets.\\ \hline
\end{tabular} 	
\caption{Input is only an abstract from CSPubSum dataset. Highlights produced by the four models are shown.  Input and author-written research highlights taken from \texttt{\url{https://www.sciencedirect.com/science/article/pii/S0010482514001565}}}	
\label{fig:sample_Abs_RHS} 
\end{figure*}

\begin{figure*}[!tbp] 
\centering 
\begin{tabular}{|p{17cm}|} \hline
{\bf Author-written research highlights:} Presenting a new two stage meta heuristic clustering algorithm based on general type 2 fuzzy sets. Incorporating a new similarity based objective function using alpha plane representation of general type 2 fuzzy sets. Implementing the proposed approach on real microarray gene expression datasets. \\\hline
{\bf PGM:} we propose a two stage sa clustering algorithm based on the concept of type 2 fcm and \colorbox{backgPm}{fcm} [UNK] utilizes a two stage sa clustering algorithm based on the new simulated annealing. Present the first hybrid meta heuristic algorithm to solve the problem. Present the first stage solution approach to solve the problem.\\\hline 
{\bf PGM + Coverage:} We present a two stage sa clustering algorithm based on the concept of planes in gt2 fss. Utilizes a two stage solution based on the new simulated annealing technology. Mechanism performs \colorbox{backgG}{very complex times} using the proposed framework. Results have been shown to be suitable for other systems. \\ \hline
{\bf PGM + SciBERT:} A two stage sa clustering algorithm is proposed based on the concept of planes in gt2 fss. The objective function of the gt2 fuzzy sa will be performed. Quality of the proposed approach is analyzed using a gt2 fuzzy cvi \colorbox{backgPm}{fuzzy cvi}. \colorbox{backgPm}{The proposed approach} is one of the first stage of the \colorbox{backgPm}{proposed approach}.\\\hline 
{\bf PGM + Coverage + SciBERT:} A two stage sa clustering algorithm based on planes in gt2 fss is proposed. Only the optimal number of clusters are obtained. The mechanism of gt2 fuzzy sa will be used to detect the optimal number of clusters. The objective function of simulated annealing and fcm is further analyzed.\\ \hline
\end{tabular}
\caption{Input is (abstract + conclusion) from CSPubSum dataset. Highlights produced by the four models are shown. Input and author-written research highlights taken from \texttt{\url{https://www.sciencedirect.com/science/article/pii/S0010482514001565}}}	
\label{fig:sample_Abs+Con_RHS} 
\end{figure*}
\begin{figure*}[!tbp] 
\centering 
\begin{tabular}{|p{17cm}|} \hline
{\bf Author-written research highlights:} Presenting a new two stage meta heuristic clustering algorithm based on general type 2 fuzzy sets. Incorporating a new similarity based objective function using alpha plane representation of general type 2 fuzzy sets. Implementing the proposed approach on real microarray gene expression datasets. \\\hline 
{\bf PGM:} A novel annealing clustering algorithm based on the sa clustering algorithm is proposed. Objective function of both planes is analyzed [UNK] on simulated annealing fcm is performed [UNK] on simulated clusters are provided.\\ \hline
{\bf PGM + Coverage:} Automated annealing clustering algorithm is proposed for the \colorbox{backgG}{stage problem of center splitting and center}. Objective function of the fcm algorithms is implemented. Optimal number of clusters is gained using the first stage. Fuzzy algorithm of a gt2 cvi on the proposed approach is presented.\\ \hline
{\bf PGM + SciBERT:} We proposed a two stage sa clustering algorithm based on the concept of planes in microarray gene expression data. We propose a two stage sa clustering framework to measure the similarity between members of each cluster. The proposed approach is primarily applied to the \colorbox{backgG}{literature}. \\\hline 
{\bf PGM + Coverage + SciBERT:} We present a two stage sa clustering approaches for microarray gene expression data. the algorithm is implemented on the concept of planes in gt2 fss. The gt2 fcm is aim to the old xie beni cvi. \\ \hline
\end{tabular}
\caption{Input is (introduction + conclusion) from CSPubSum dataset. Highlights produced by the four models are shown. Input and author-written research highlights taken from \texttt{\url{https://www.sciencedirect.com/science/article/pii/S0010482514001565}}}	
\label{fig:sample_Int+Con_RHS} 
\end{figure*}

\subsection{Case study on MixSub dataset}
We now present an example demonstrating the output produced by the four variants of pointer-generator model for the MixSub dataset. Figure \ref{fig:sample_Abs_RHS_extended_dataset} displays the outputs when the models take only the abstract as the input. We observe that the last two sentences produced by the plain pointer-generator model are identical. This issue goes away when coverage is added but now the model's output is not entirely factually correct: it generates ``The reflection of the plasma photonic crystal to near infrared radiation \textit{increases} with the wave angle.'' while the abstract mentions ``The reflection of the plasma photonic crystal to near infrared radiation \textit{decreases} with increasing of the incident wave angle ...''. This factual error disappears when SciBERT is added. The output quality improves further when coverage is added. 
While PGM with SciBERT (but without coverage) outputs ``the near infrared filter photonic crystal is adjusted'', the final model (PGM + coverage + SciBERT) is more precise: ``\textit{infrared radiation pass band} can be adjusted''. Therefore, the highlights produced by \textbf{(PGM + Coverage + SciBERT)} seem to be most satisfactory though its last sentence has a syntax error due to a missing relative pronoun: ``A transfer method is proposed for infrared radiation pass band [which] can be adjusted ...''.  

\begin{figure*}[!tbp] 
\centering 
\begin{tabular}{ |p{17cm}|} \hline
{\bf Author-written research highlights:} A plasma photonic crystal composed of ito and plasma is proposed. The performance of ppc in near infrared radiation modulation is researched by tmm. The near infrared radiation pass band can be adjusted by plasma frequency. The ppc has a potential application in tunable near infrared filter devices.\\\hline
{\bf PGM:} A plasma photonic crystal for infrared radiation modulation was proposed. The plasma photonic crystal was researched by the changing changing of plasma frequency of plasma. The incident wave angles have little effect on the transmission of plasma. \colorbox{backgPm}{The incident wave angles have little effect on the} \colorbox{backgPm}{transmission of plasma.}\\\hline 	    
{\bf PGM + Coverage:} A plasma photonic crystal for infrared radiation and plasma is proposed. The incidence wave angles can be adjusted by the changing plasma oxide band. The incident wave angles have little effect on the transmission of plasma photonic crystal. The reflection of the plasma photonic crystal to near infrared radiation \colorbox{backgG}{increases with the wave angle}. \\ \hline
{\bf PGM + SciBERT:} A plasma photonic crystal for infrared radiation modulation is proposed.
The near infrared filter photonic crystal is adjusted by the changing of plasma frequency of plasma photonic crystal in near infrared filter devices. The proposed plasma photonic crystal has a potential application in tunable near infrared filter devices. \\\hline
{\bf PGM + Coverage + SciBERT:} A plasma photonic crystal for infrared radiation modulation is proposed.
A transfer matrix method is proposed for infrared radiation pass band can be adjusted by the changing of plasma frequency.
The proposed plasma photonic crystal has a potential application in tunable near infrared filter devices.\\ \hline
\end{tabular} 	
\caption{Input is only the abstract of an article from the MixSub dataset. Highlights produced by the four models are shown. Input and author-written research highlights taken from \texttt{\url{https://www.sciencedirect.com/science/article/pii/S1567173920301292}}}	
\label{fig:sample_Abs_RHS_extended_dataset} 
\end{figure*}

\section{Conclusion and Future Work}
\label{Conclusion}
We applied four different deep neural models to generate research highlights from a research paper. We experimented with different input types for each model for one of the datasets while we used only the abstract as input for the other dataset. The pointer-generator model with SciBERT and coverage mechanism achieved the best performance in each case. But the predicted research highlights are not yet perfect in terms of syntax and semantics. We are currently exploring other techniques to address these issues. A few other research directions  would be to generate highlights that summarize a set of related papers, and to build a database containing research findings from different papers with links connecting semantically-related findings.  

\section*{Declarations}
\subsection*{Availability of supporting data}
The dataset CSPubSum is publicly available\footnote{\url{https://github.com/EdCo95/scientific-paper-summarisation/tree/master/DataDownloader}}. The MixSub dataset used in the paper has been publicly released by us on github\footnote{\url{https://github.com/tohidarehman/Highlights-Generation-ResearchPaper}}. 
\subsection*{Declaration of competing interests}
The authors declare that they have no competing interests.
\subsection*{Funding}
The second author is partially supported by the Department of Science and Technology, Government of India under grant CRG/2021/000803. 
\bibliographystyle{unsrt}
\bibliography{pgm}

\begin{IEEEbiography}[{\includegraphics[width=1in,height=1.25in,clip,keepaspectratio]{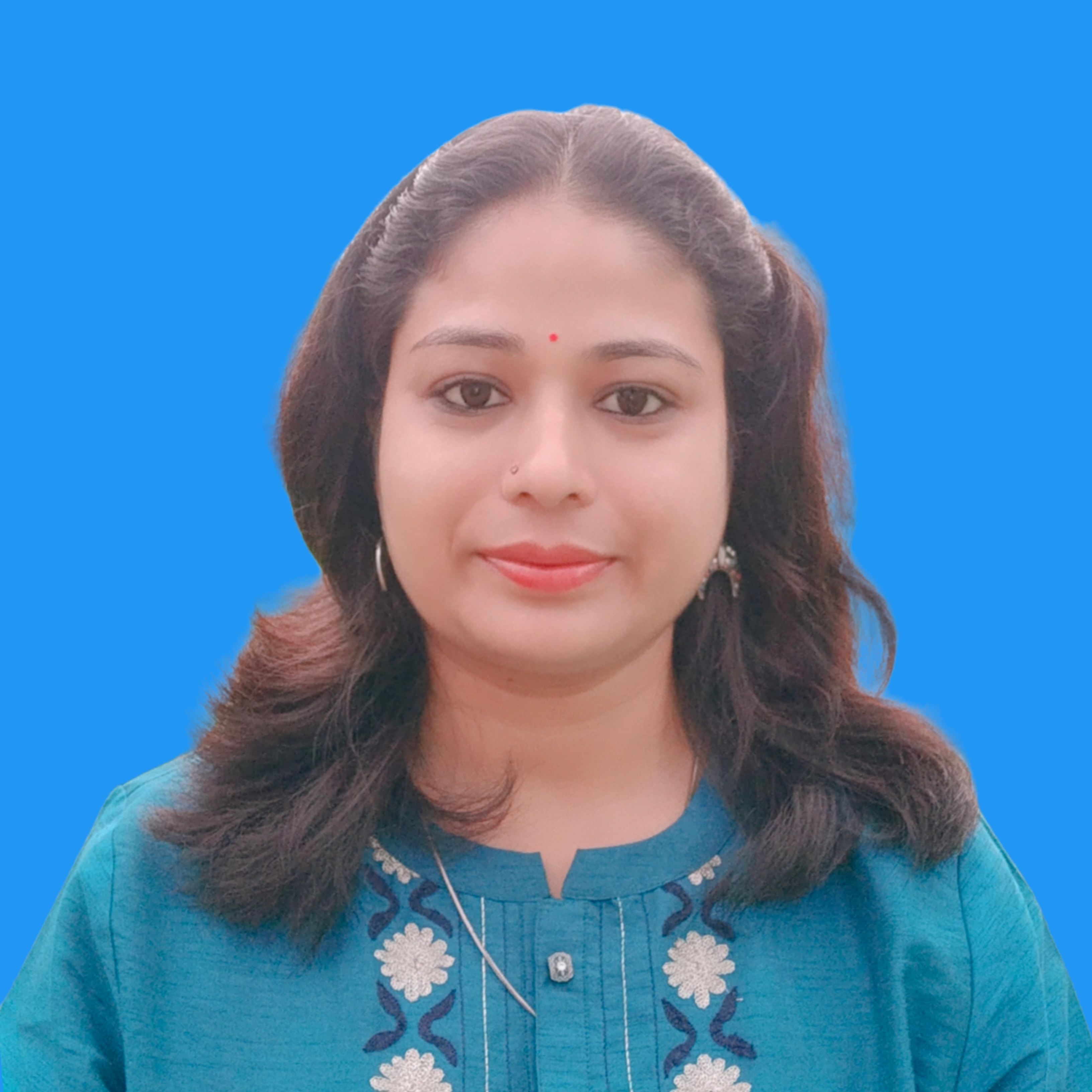}}]{Tohida Rehman} is an Assistant Professor with the Department of Information Technology,
Jadavpur University, India. From 2014 to 2018, she was an Assistant Professor with the Department of Computer Science, Surendranath College, Calcutta University, India. She has more than nine
years of teaching experience. Her current research interests include machine learning and natural language processing. Her current work focuses on improving text summarization.
\end{IEEEbiography}
\vskip -2\baselineskip plus -1fil
\begin{IEEEbiography}[{\includegraphics[width=1in,height=1.25in,clip,keepaspectratio]{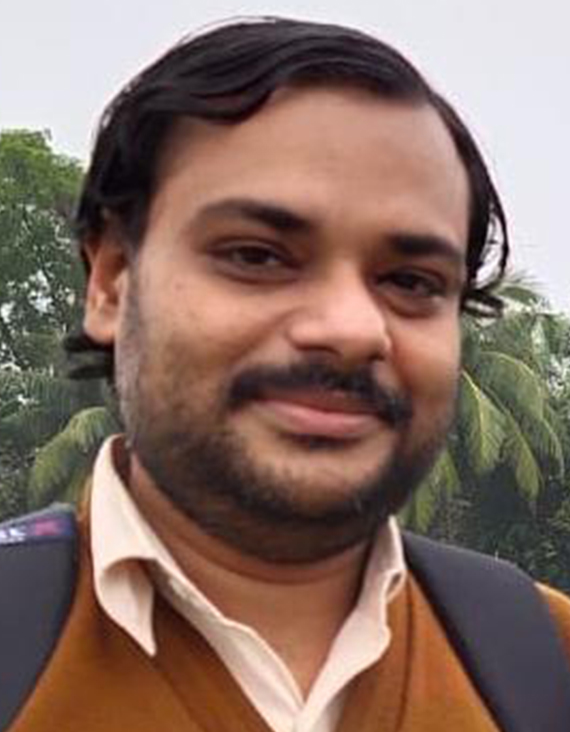}}]{Debarshi Kumar Sanyal} received the B.E. degree in information technology and the Ph.D. degree in engineering from Jadavpur University,
Kolkata, in 2005 and 2012, respectively. He is an Assistant Professor with the School of Mathematical and Computational Sciences, Indian Association for the Cultivation of Science, Kolkata, India.
He was with IIT Kharagpur, KIIT Deemed University, Xilinx India Pvt. Ltd., Interra Systems India
Pvt. Ltd., and Infosys Ltd. His current research interests include natural language processing, digital library technologies, information retrieval, and machine learning.
\end{IEEEbiography}
\vskip -2\baselineskip plus -1fil
\begin{IEEEbiography}[{\includegraphics[width=1in,height=1.25in, clip,keepaspectratio]{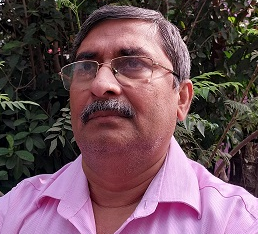}}] {SAMIRAN CHATTOPADHYAY} is a Pro Vice Chancellor in Techno India University, West Bengal and a former professor of the department of Information Technology, Jadavpur University. He works in the areas of machine intelligence and its applications, wireless networks, and network security. His current work mostly centers around the design and creation of computer models for decision analysis and optimization, particularly in the fields of HCI, high throughput wireless networks, healthcare, power engineering, and technology-enabled learning. He has more than three decades of experience in teaching and research in the broad domain of computer science and engineering (CSE). He is the author of about 180 articles in reputed journals and conference proceedings including more than 60 high impact journal publications.
\end{IEEEbiography}
\vskip -2\baselineskip plus -1fil
\begin{IEEEbiography}[{\includegraphics[width=1in,height=1.25in,clip,keepaspectratio]{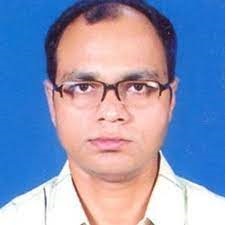}}]{Plaban Kumar Bhowmick}  received the M.S. and Ph.D. degrees from the Department of Computer Science and Engineering, IIT Kharagpur. He is an Assistant Professor with the G. S. Sanyal School of Telecommunications, Centre of Excellence in Artificial Intelligence, Indian Institute of Technology Kharagpur. His research interests include automated answer grading, augmenting learner experience, and graph machine learning. 
\end{IEEEbiography}

\vskip -2\baselineskip plus -1fil
\begin{IEEEbiography}[{\includegraphics[width=1in,height=1.25in,clip,keepaspectratio]{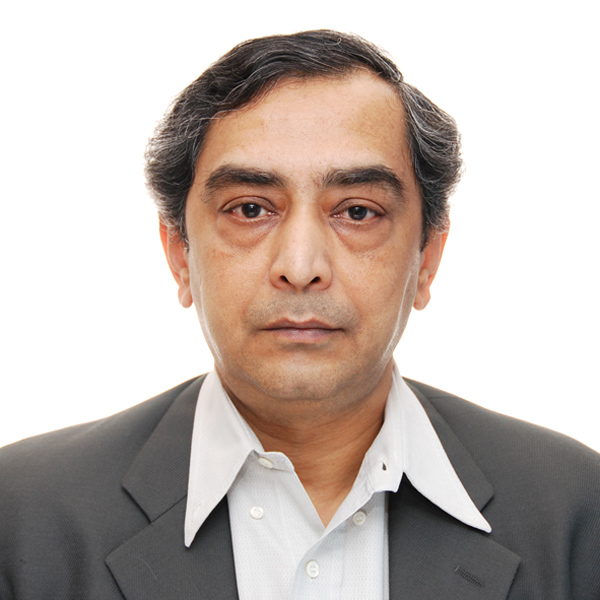}}]{Partha Pratim Das} (Member, IEEE) received the B.Tech., M.Tech., and Ph.D. degrees from the Department of Electronics and Electrical Communication, IIT Kharagpur, in 1984, 1985, and, 1988 respectively. 

He is a Visiting Professor with the Department of Computer Science, Ashoka University. He is on leave from IIT Kharagpur, where he was a Professor with the Department of Computer Science and Engineering. He has over 22 years of experience in teaching and research with IIT Kharagpur, and about 13 years of experience in software industry, including start-ups. Over the past ten years, he has led the Development of National Digital Library of India (NDLI) Project, MoE, GoI, as a Joint Principal Investigator. He has also developed a unique vertical DEEPAK: Disability Education and Engagement Portal for Access to Knowledge. He has also led engineering entrepreneurship education, research, facilitation, and deployment with IIT Kharagpur, from 2013 to 2020. He is a strong proponent of online education. He has been a Key Instructor of three courses with SWAYAM-NPTEL, since 2016. During pandemic, he has instrumental in making learning material available to the students through NDLI. He currently works on the following problems: hands-free control and immersive navigation of Chandrayaan and Mangalyaan images on large displays (with ISRO), smart knowledge transfer for legacy software projects, automated interpretation of Bharatanatyam dance, and the development of Indian food atlas and food knowledge graph. 

Dr. Das received the Young Scientist/Engineer Award from the Indian National Science Academy, in 1990, the Indian National Academy of Engineering, in 1996, and the Indian Academy of Sciences, in 1992. Being in the leadership team of NDLI, he was recognized for his contributions toward online education during pandemic through several awards, including the OE Awards for Excellence: Open Resilience, in 2020; the SM4E Award: Innovation@COVID-19, in 2021; and the World Summit Award: Learning and Education, in 2021.
\end{IEEEbiography}
\EOD

\end{document}